\PassOptionsToPackage{table}{xcolor}
\documentclass[]{bytedance_seed}
\usepackage{iftex}
\usepackage{booktabs}
\usepackage{tabularx}
\usepackage{multirow}
\usepackage{graphicx}
\usepackage{pifont}
\usepackage{colortbl}
\usepackage{amssymb}

\providecommand{\cmark}{\ding{51}}
\providecommand{\xmark}{\ding{55}}

\definecolor{taskgray}{gray}{0.92}

\usepackage{amsmath,amsfonts,bm}

\def\eqref#1{equation~\ref{#1}}

\def\1{\bm{1}}

\DeclareMathAlphabet{\mathsfit}{\encodingdefault}{\sfdefault}{m}{sl}
\SetMathAlphabet{\mathsfit}{bold}{\encodingdefault}{\sfdefault}{bx}{n}

\usepackage{url}
\usepackage{enumitem}

\newsavebox{\templatetablebox}
\let\templatetabular\tabular
\let\endtemplatetabular\endtabular
\renewenvironment{tabular}{%
  \begin{lrbox}{\templatetablebox}\templatetabular
}{%
  \endtemplatetabular\end{lrbox}%
  \ifdim\wd\templatetablebox>\linewidth
    \resizebox{\linewidth}{!}{\usebox{\templatetablebox}}%
  \else
    \usebox{\templatetablebox}%
  \fi
}
\AtBeginEnvironment{quote}{\raggedright}

\newenvironment{tasklist}{%
  \begin{list}{\textbullet}{%
    \setlength{\leftmargin}{1.2em}%
    \setlength{\labelwidth}{0.7em}%
    \setlength{\labelsep}{0.5em}%
    \setlength{\topsep}{3pt}%
    \setlength{\itemsep}{2pt}
    \setlength{\parsep}{0pt}%
    \setlength{\partopsep}{0pt}%
  }%
}{\end{list}}

\newcommand{\Yijia}[1]{\textcolor{blue}{[Yijia: #1]}}

\usepackage{placeins}
\usepackage{environ}
\usepackage{adjustbox}
\let\originalequation\equation
\let\endoriginalequation\endequation
\RenewEnviron{equation}{%
  \originalequation
  \begin{adjustbox}{max width=0.90\linewidth}
    $\displaystyle\BODY$
  \end{adjustbox}
  \endoriginalequation
}

\title{ALIVE: Making Streaming Video Models Interactive}
\author[1,3,*]{Jinzhao Li}
\author[2,3,*]{Yuan Xu}
\author[4]{Jinxuan Wu}
\author[1]{Yukun Wang}
\author[1]{\\Yijia Lei}
\author[2]{Hanyu Ji}
\author[1]{Yinuo Chen}
\author[1]{Zining Huang}
\author[1]{\\Chi Zhang}
\author[2]{Hang Ye}
\author[2]{Yiwu Zhong}
\author[3]{\\Honglei Yan}
\author[3]{Panwang Pan}
\author[2]{Yizhou Wang}
\author[5]{Yin Li}
\author[1,\dagger]{Miao Liu}

\affiliation[1]{Tsinghua University}
\affiliation[2]{Peking University}
\affiliation[3]{ByteDance}
\makeatletter
\g@addto@macro\affiliationlist{\\\affiliationformat[4]{Beihang University}}
\makeatother
\affiliation[5]{University of Wisconsin\textendash{}Madison}

\contribution[*]{Equal contribution}
\contribution[\dagger]{Corresponding author}
\makeatletter
\g@addto@macro\contributionlist{\par{\contributionfont
  \email{lijinzha22@mails.tsinghua.edu.cn}\quad
  \email{miaoliu@mail.tsinghua.edu.cn}}}
\makeatother

\abstract{Streaming video LLMs can continuously observe the world, yet often handle user requests in isolation. We introduce \textbf{ALIVE}, an interactive streaming video model supporting proactive monitoring, streaming task coordination, and feedback-seeking interaction. We construct \textbf{ALIVE-SFT-288K} and \textbf{ALIVE-RL-3K} for two-stage training. Contrastive supervised fine-tuning teaches when and how to respond, while reinforcement learning with \emph{Behavior-Aligned Credit Assignment} (BACA) combines trajectory-level and local advantages to refine decisions, responses, and silence across multi-turn interactions. We also introduce \textbf{FSI-Bench} to evaluate feedback seeking and subsequent task completion. ALIVE achieves the highest overall scores among the compared models on four public benchmarks and FSI-Bench, demonstrating strong streaming understanding and interactive capabilities.

}
\checkdata[Project Page]{\href{https://lijinzhao30.github.io/ALIVE/}{https://lijinzhao30.github.io/ALIVE/}}
\checkdata[Code]{\href{https://github.com/lijinzhao30/ALIVE}{https://github.com/lijinzhao30/ALIVE}}
\begin{document}
\maketitle
\begin{figure}[t]
\input{secs/teaser}
\end{figure}
\section{Introduction}


Advances in large language models (LLMs) are pushing conversational AI beyond multi-turn dialogues toward interactive assistants that can respond to diverse user requests, manage multiple concurrent tasks, and, when appropriate, proactively engage users as those tasks evolve~\citep{ye2025realwebassistbenchmarklonghorizonweb,lu2024proactiveagentshiftingllm}. 
Bringing these capabilities into the physical world points toward a long-standing vision of multimodal interactive AI~\citep{driess2023palm,Li_2026_CVPR,huang2022inner}: systems such as \textit{JARVIS in Iron Man} that continuously perceive streams of inputs while seamlessly interacting with human users, augmenting our ability to understand and act in complex, dynamic environments.

Streaming video LLMs lie at the very heart of this vision. Recent models~\citep{chen2024videollmonlineonlinevideolarge,xiong2025streamingvideounderstandingmultiround,xu2026streamingvlmrealtimeunderstandinginfinite} can continuously process incoming video, reason about past observations~\citep{zeng2025streamforest}, describe ongoing activities~\citep{Chen_2025_CVPR,wu2024videollm}, and monitor events as they occur. Yet today's streaming video LLMs remain largely passive, often responding to individual requests in isolation. They lack a critical capability of interactive assistants: the ability to maintain and revise user tasks over time and proactively engage users as the video unfolds. This gap in interactivity substantially limits their potential as a foundation for multimodal interactive AI.

In this work, we aim to transform streaming video LLMs from passive observers into interactive assistants. Given that modern video LLMs largely inherit the architecture of LLMs~\citep{li2025videochat,lin2024video}, we ask a key question: \textit{What data representation and learning framework can endow existing video LLMs with interactive capabilities?}

Addressing this question poses two challenges. First, constructing streaming interaction data requires aligning user requests, feedback, and assistant responses with the visual evidence available at each moment. Second, models must learn when and how to respond as context and task states evolve across multiple turns, where similar observations can call for different behaviors.

To address these challenges, we first identify a set of core capabilities required for interactive streaming video understanding and curate ALIVE-SFT-288K and ALIVE-RL-3K for two-stage post-training. We then develop a learning framework that explicitly supports these capabilities. Contrastive SFT teaches strategy selection and response gating through controlled variations in interaction conditions and available evidence. Reinforcement learning with \emph{Behavior-Aligned Credit Assignment} (BACA) further refines these behaviors by combining trajectory-level advantages with local advantages for strategy selection, task response, and critical silence. We also establish FSI-Bench to evaluate feedback-seeking interaction in streaming video LLMs.

Our experiments show that an offline video LLM can indeed be transformed through post-training into a capable interactive streaming video LLM. The resulting model preserves its ability to understand a streaming video while acquiring substantially richer interactive behaviors: it can simultaneously manage multiple user-specified tasks, modify or cancel existing tasks, and proactively seek user feedback based on prior requests and events unfolding in the video. These results suggest a path toward streaming video LLMs that do not merely watch and respond, but sustain ongoing interactions with users as the video streams. Our \textbf{contributions} are summarized as follows:
\begin{itemize}[leftmargin=1em, itemsep=2pt, topsep=2pt]
    \item We formulate \textbf{interactive streaming video understanding} as a set of core capabilities, including proactive monitoring, streaming task coordination, and feedback-seeking interaction. Building on this formulation, we introduce \textbf{ALIVE-SFT-288K} and \textbf{ALIVE-RL-3K} for two-stage post-training, together with \textbf{FSI-Bench}, a benchmark for evaluating feedback-seeking interaction.\vspace{-0.3em}
    \item We introduce \textbf{ALIVE}, an interactive streaming video model trained through \textbf{contrastive SFT} and \textbf{reinforcement learning with BACA}.\vspace{-0.3em}
    \item We show that ALIVE achieves highly competitive performance across multiple streaming video benchmarks, preserving strong video understanding capabilities while substantially improving performance on interactive tasks.
\end{itemize}

\section{Related Work}

\noindent\textbf{Streaming Video Models.}
Streaming video models process continuously arriving observations~\citep{liu2024streamchat,huang2025online} and determine when to respond~\citep{wang2025videollm}. VideoLLM-online couples temporally aligned dialogue with continuous inference~\citep{chen2024videollmonlineonlinevideolarge}, while Dispider separates perception, decision, and reaction for asynchronous interaction~\citep{qian2025dispiderenablingvideollms}. StreamBridge and AURA address sustained observation through memory and context management~\citep{wang2025streambridgeturningofflinevideo,lu2026auraalwaysonunderstandingrealtime}. JoyAI-VL-Interaction integrates silence, response, and delegation with an asynchronous background model~\citep{yao2026joyaivlinteractionrealtimevisionlanguageinteraction}. ALIVE focuses on maintaining evolving user requests and resolving missing prerequisites through clarification, user execution, or continued monitoring.

\noindent\textbf{Post-training for Streaming Video Understanding.}
Streaming SFT learns response content and timing from temporally aligned dialogues, with supervision of response states and balancing of silence and speech~\citep{chen2024videollmonlineonlinevideolarge,xia2026streamingvideoinstructiontuning,lu2026auraalwaysonunderstandingrealtime}. Our contrastive SFT further supervises how interaction conditions and available evidence change the appropriate strategy and response gate. Beyond SFT, PPO~\citep{schulman2017proximal} and GRPO~\citep{shao2024deepseekmath} provide policy-optimization frameworks for LLM post-training. GiGPO~\citep{feng2025groupingroup} further addresses multi-turn credit assignment through episode-level and step-level advantages. In streaming video, MMDuet2 uses RL to improve response timing and correctness while penalizing repetition~\citep{wang2025mmduet2enhancingproactiveinteraction}. StreamPro uses GRPO with aggregated turn-level and trajectory-level rewards~\citep{li2026streampro}. Our BACA combines trajectory-level advantages with local advantages over behavioral units for strategy selection, task response, and critical silence.

\noindent\textbf{Streaming Video Datasets and Benchmarks.}
Streamo-Instruct covers diverse streaming tasks, while ProAssist provides proactive assistant--user dialogues grounded in egocentric videos~\citep{xia2026streamingvideoinstructiontuning,zhang2025proactive}. StreamingBench and OVO-Bench evaluate online perception and temporal understanding~\citep{lin2024streamingbenchassessinggapmllms,li2025ovobenchfarvideollmsrealworld}. ProactiveVideoQA focuses on proactive response timing and quality, while IPIBench covers task updates and interleaved requests~\citep{wang2025proactivevideoqacomprehensivebenchmarkevaluating,li2026ipibench}. Table~\ref{tab:streaming_task_comparison} compares training datasets. ALIVE-SFT-288K and ALIVE-RL-3K support two-stage training across monitoring, task coordination, and feedback seeking. FSI-Bench evaluates strategy selection and subsequent task completion when additional information, user action, or observation is required.

\begin{table}[t]
  \centering
  \caption{Comparison of streaming video training datasets.
    \textbf{PM}: Proactive Monitoring; \textbf{STC}: Streaming Task Coordination;
    \textbf{FSI}: Feedback-Seeking Interaction.
    \cmark/\xmark: explicit/no explicit support.
    \textbf{Stage}: supervised fine-tuning (SFT) or reinforcement learning (RL).
    Scale follows each work's counting unit (K = thousand).}
  \label{tab:streaming_task_comparison}
  \small
  \renewcommand{\textbf}[1]{{\sffamily\bfseries #1}}
  \renewcommand{\arraystretch}{1.12}
  \setlength{\tabcolsep}{4pt}
  \let\tabular\templatetabular
  \let\endtabular\endtemplatetabular
    \begin{tabularx}{\linewidth}{@{}>{\hsize=2.4\hsize\raggedright\arraybackslash}X*{7}{>{\hsize=.8\hsize\centering\arraybackslash}X}@{}}
      \toprule
      \textbf{Dataset}
      & \shortstack{\textbf{Multi-}\\\textbf{turn}}
      & \shortstack{\textbf{Visual}\\\textbf{source}}
      & \textbf{Scale}
      & \textbf{PM}
      & \textbf{STC}
      & \textbf{FSI}
      & \textbf{Stage} \\
      \midrule
      LIVE~\cite{chen2024videollmonlineonlinevideolarge}
      & \cmark & Ego & 134K & \cmark & \xmark & \xmark & SFT \\
      MMDuet2-SFT~\cite{wang2025mmduet2enhancingproactiveinteraction}
      & \cmark & Both & 50K & \cmark & \xmark & \xmark & SFT \\
      Streamo-Instruct~\cite{xia2026streamingvideoinstructiontuning}
      & \xmark & Both & 465K & \cmark & \xmark & \xmark & SFT \\
      VideoChat3-OL617K~\cite{li2026videochat3}
      & \xmark & Both & 617K & \cmark & \xmark & \xmark & SFT \\
      ProAssist~\cite{zhang2025proactive}
      & \cmark & Ego & 28K & \cmark & \xmark & \xmark & SFT \\
      StreamPro-SFT-63K~\cite{li2026streampro}
      & \xmark & Both & 63K & \cmark & \xmark & \xmark & SFT \\
      \rowcolor{taskgray}
      \textbf{ALIVE-SFT-288K}
      & \cmark & Both & 288K & \cmark & \cmark & \cmark & SFT \\
      \midrule
      MMDuet2-RL~\cite{wang2025mmduet2enhancingproactiveinteraction}
      & \cmark & Both & 1.9K & \cmark & \xmark & \xmark & RL \\
      StreamPro-RL-3K~\cite{li2026streampro}
      & \xmark & Both & 3K & \cmark & \xmark & \xmark & RL \\
      VST-RL~\cite{guan2026video}
      & \xmark & Both & 11K & \xmark & \xmark & \xmark & RL \\
      \rowcolor{taskgray}
      \textbf{ALIVE-RL-3K}
      & \cmark & Both & 3K & \cmark & \cmark & \cmark & RL \\
      \bottomrule
    \end{tabularx}
  \vspace{-1em}
\end{table}

\section{Streaming Interactivity: Problem Formulation and Datasets}
\label{sec:data}

\subsection{Problem Formulation}
We consider a streaming video model that continuously receives video observations and user utterances. At decision time $t$, its output depends only on the observed video prefix $V_{1:t}$ and the interaction history $H_t$, comprising received user utterances and prior model outputs. All mappings below describe causal rollouts: each output uses only information available at its own decision time.

A \textbf{passive streaming} model executes a fixed query $Q_{t_0} \in H_t$ issued at time $t_0$:
\begin{equation}
    f_\theta(Q_{t_0}, V_{1:t})
    \rightarrow (r_{t_0},\ldots,r_t),
\label{eq:passive_streaming}
\end{equation}
where $(r_{t_0},\ldots,r_t)$ are responses from $t_0$ to $t$ including the choice of remaining silence (\texttt{SILENCE}). This formulation covers immediate question answering and delayed or repeated event monitoring. Here, \emph{passive} refers to executing a fixed request without accommodating new requests or seeking additional user input, not to the absence of proactive responses.

An \textbf{interactive streaming} model retains the same streaming capabilities as the passive setting, while supporting a broader range of outputs and evolving interactions. Before receiving further user input, its rollout is similar yet with richer output
\begin{equation}
    f_\theta(Q_{t_0}, V_{1:t})
    \rightarrow (o_{t_0},\ldots,o_t),
\ \ \text{with}\ \ 
    o_\tau =
    \begin{cases}
        r_\tau, & \text{respond or remain silent},\\
        c_\tau, & \text{request clarification or feedback},\\
        a_\tau, & \text{request user execution}.
    \end{cases}
\label{eq:interactive_outputs}
\end{equation}

Unlike passive streaming, interactive streaming maintains an evolving interaction history $H_t$. At a later decision time $\tau>t$, a new interaction $\Delta H_\tau$ updates this state:
\begin{equation}
    H_\tau
    =
    \operatorname{Update}(H_t, \Delta H_\tau),
\end{equation}
where $\Delta H_\tau$ may correspond to a new user query that adds, cancels, or modifies a request, user feedback that resolves a previous clarification request $c_t$, or newly observed evidence resulting from an execution request $a_t$. With the updated state and retained interaction history, the model continues with a similar streaming processing
\begin{equation}
    f_\theta(H_\tau, V_{1:\tau'})
    \rightarrow O_{\tau:\tau'},
    \qquad \tau' \geq \tau.
\label{eq:interactive_continue}
\end{equation}
This process unfolds along a shared arrow of time, where both the user and model can initiate interaction, update one another, and continue acting on the evolving video stream. Interactive streaming thus enables bidirectional communication within a single continuous trajectory.


\subsection{Interactive Streaming Task Taxonomy}
\label{sec:task_definition}

As aforementioned, passive streaming models primarily focus on proactive monitoring, while interactive streaming further extends to task coordination and feedback seeking. Accordingly, we briefly introduce the core interactive streaming capabilities across three complementary families.

\noindent\textbf{Proactive Monitoring}.\ The user specifies a condition to monitor, and the model observes the stream until sufficient evidence supports a response:
\begin{tasklist}
    \item \textit{Proactive Timing}: The model issues a notification when the target condition is first verifiably satisfied, completing the task.
    \item \textit{Proactive Understanding}: The model answers an associated visual question once both the trigger and sufficient answer evidence are available.
    \item \textit{Repeated Proactiveness}: The model maintains the request across multiple occurrences and responds once for each distinct occurrence.
\end{tasklist}

\noindent\textbf{Streaming Task Coordination}.\ The model maintains and updates active commitments as user requests evolve or coexist:
\begin{tasklist}
    \item \textit{Task State Update}: The user cancels or modifies an existing monitoring task, and the model updates subsequent behavior accordingly.
    \item \textit{Concurrent Requests}: The model handles multiple monitoring objectives or interleaved immediate questions while preserving all remaining active commitments.
\end{tasklist}

\noindent\textbf{Feedback-Seeking Interaction}.\ When the current context is insufficient to resolve a request, the model identifies the missing prerequisite and actively seeks the required feedback:
\begin{tasklist}
    \item \textit{Ask-for-Monitor}: When the answer depends on future observations, the model asks whether the user wants to establish a monitoring task and responds once the relevant evidence appears.
    \item \textit{Ask-for-Information}: When the request is ambiguous, the model asks for clarification and uses the additional information to resolve the original query.
    \item \textit{Ask-for-Execution}: When necessary visual evidence is unavailable, the model requests a concrete user action that reveals the missing information before answering.
\end{tasklist}

Across these tasks, the model must preserve valid commitments and incorporate subsequent user feedback or visual evidence into its ongoing interaction state.

\subsection{Data Construction and Datasets}
\label{sec:data_construction}

\begin{figure*}[t]
    \centering
    \includegraphics[width=\textwidth]{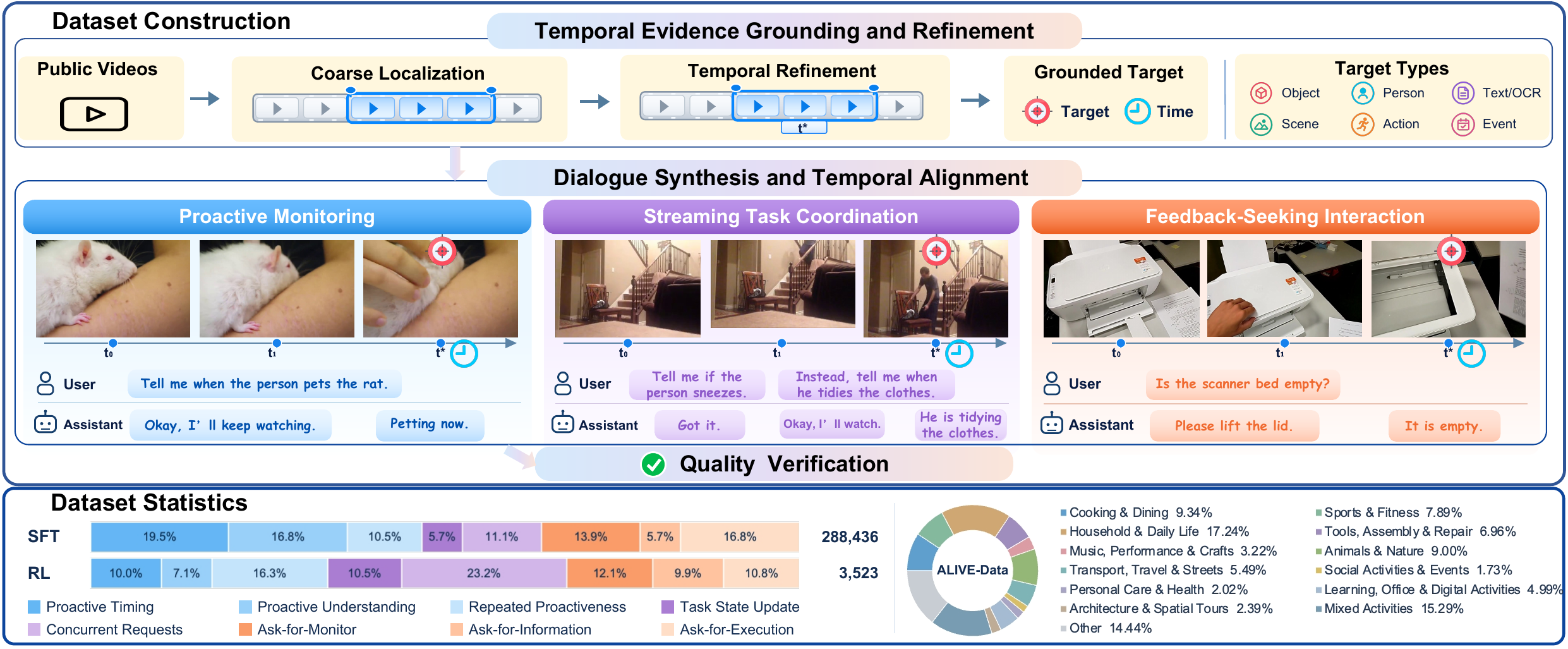}
    \caption{\textbf{Construction and composition of ALIVE training data.} \textbf{Top:} Targets in public videos are coarsely localized and temporally refined. Grounded evidence guides dialogue synthesis and alignment for proactive monitoring, streaming task coordination, and feedback-seeking interaction, followed by quality verification. \textbf{Bottom:} Task distributions are shown separately for SFT and RL. The donut chart summarizes video-content categories jointly over both training sets.}
    \label{fig:dataset}\label{fig:statistics}\vspace{-1.5em}
\end{figure*}


We construct ALIVE-SFT-288K and ALIVE-RL-3K using public video segments and annotations from diverse sources. These include TimeLens2~\citep{zhu2026timelens2generalistvideotemporal}, LLaVA-Video~\citep{zhang2024llava}, TimeLens~\citep{zhang2026timelensrethinkingvideotemporal}, Shot2Story~\citep{han2025shot2storynewbenchmarkcomprehensive}, VideoGPT+~\citep{maaz2024videogptintegratingimagevideo}, VideoRefer-700K~\citep{yuan2025videorefersuiteadvancingspatialtemporal}, QVHighlights~\citep{lei2021qvhighlightsdetectingmomentshighlights}, Vript~\citep{yang2024vriptvideoworththousands}, WebVid~\citep{bain2022frozentimejointvideo}, DiDeMo~\citep{hendricks2017localizingmomentsvideonatural}, Ref-YouTube-VOS~\citep{seo2020urvos}, Nymeria~\citep{ma2024nymeriamassivecollectionmultimodal}, EPIC-KITCHENS~\citep{damen2020epickitchensdatasetcollectionchallenges}, Charades-Ego~\citep{sigurdsson2018charadesegolargescaledatasetpaired}, Ego4D~\citep{grauman2022ego4d}, HoloAssist~\citep{wang2023holoassistegocentrichumaninteraction}, Ego-Exo4D~\citep{grauman2024egoexo4dunderstandingskilledhuman}, DenseStep2M~\citep{ge2026densestep2mscalabletrainingfreepipeline}, COIN~\citep{tang2019coinlargescaledatasetcomprehensive}, HACS~\citep{zhao2019hacshumanactionclips}, Charades-STA~\citep{gao2017talltemporalactivitylocalization}, STAR~\citep{wu2024star}, ActivityNet~\citep{Heilbron_2015_CVPR}, and Oops~\citep{epstein2019oopspredictingunintentionalaction}.

We pair these video segments with timestamped user requests, feedback, and assistant responses or silence to construct interaction trajectories. As illustrated in Fig.~\ref{fig:dataset}, construction proceeds through temporal evidence grounding, dialogue synthesis and alignment, and quality verification.

\medskip
\noindent\textbf{Temporal evidence grounding and refinement.}
For videos with existing descriptions and approximate timestamps, we use them to identify candidate intervals; otherwise, an offline MLLM first detects relevant targets and coarsely localizes their occurrence. We then refine each candidate using an offline MLLM with access to video context both before and after the candidate event. This bidirectional temporal context helps identify the earliest sampled time at which the required visual condition is met, such as entity recognizability, text readability, scene visibility, or action completion.

\noindent\textbf{Dialogue synthesis and temporal alignment.}
Given grounded targets and timestamps, we synthesize dialogues that pair user requests with appropriate assistant responses and, where needed, task updates or feedback exchanges. We align these exchanges with the evolving task state and visual evidence, so that each answer follows the observations or user feedback needed to support it.

\noindent\textbf{Quality verification.}
We verify two aspects of each example. Interaction-strategy validity: the selected strategy must be appropriate for the current task state and information gap, and any requested clarification, action, or continued observation must be relevant to resolving the request. Causal answerability: each answer must be supported by the video observed up to its assigned timestamp and the preceding interaction history, without relying on future frames or subsequent user feedback. Examples that fail either check are filtered.

\noindent\textbf{Statistics.}
Through this construction process, we generate ALIVE-SFT-288K and ALIVE-RL-3K, comprising 288,436 and 3,523 interaction examples for supervised fine-tuning and reinforcement learning, respectively. Both datasets span diverse scenarios and cover eight task types across proactive monitoring, streaming task coordination, and feedback-seeking interaction. Fig.~\ref{fig:dataset} summarizes the task distributions and video-content categories.

\section{Learning Interactive Streaming Video LLMs}
\label{sec:method}

Streaming interaction requires a model to select an appropriate interaction strategy, respond accurately when the necessary evidence becomes available, and remain silent when a response is unwarranted. These behaviors depend jointly on visual observations and the evolving interaction history. We train ALIVE in two stages. Contrastive supervised fine-tuning establishes these conditional interaction behaviors, while reinforcement learning further optimizes them on model-generated trajectories through \emph{Behavior-Aligned Credit Assignment}.

\noindent\textbf{Output format.}
\label{sec:formulation}
At each decision time $t$, the model generates $o_t\sim p_\theta(\cdot\mid V_{1:t},H_t)$ from the observed video prefix and interaction history. A control token indicates whether to respond ($R$) or remain silent ($S$). When responding, the subsequent text expresses the appropriate strategy, such as answering directly, proposing continued monitoring, requesting clarification, or asking the user to perform an action. All outputs are generated by a single autoregressive model.

\subsection{Contrastive Supervised Fine-Tuning}
\label{sec:sft}

The appropriate interaction behavior is not determined by visual content alone. The same observation may warrant different outputs depending on the active request, task status, or available user feedback. We construct two complementary families of controlled contrasts to supervise these dependencies.

\noindent\textbf{Response-gate contrasts.}
\label{sec:response_gating}
We hold the visual prefix and comparison time fixed while varying an interaction condition that changes whether the model should respond. These conditions include query relevance, task status, and feedback state. For example, the same target appearance requires a notification under an active monitoring request but silence after cancellation or fulfillment of a single-occurrence request. These contrasts provide explicit supervision for warranted responses and critical silence under visually similar conditions.

\noindent\textbf{Action-strategy contrasts.}
\label{sec:action_strategy}
We vary evidence availability or referential ambiguity while preserving the underlying task. When sufficient evidence is available, the model should answer directly. An ambiguous reference requires clarification, whereas missing visual evidence may require a concrete user action or a proposal to continue observing. We pair these situations with their direct-answer counterparts to teach strategy selection according to the information available at the time. Both sides use the response gate, with the distinction learned through supervision on the response text.

\begin{figure*}[t]
    \centering
    \includegraphics[width=0.95\textwidth]{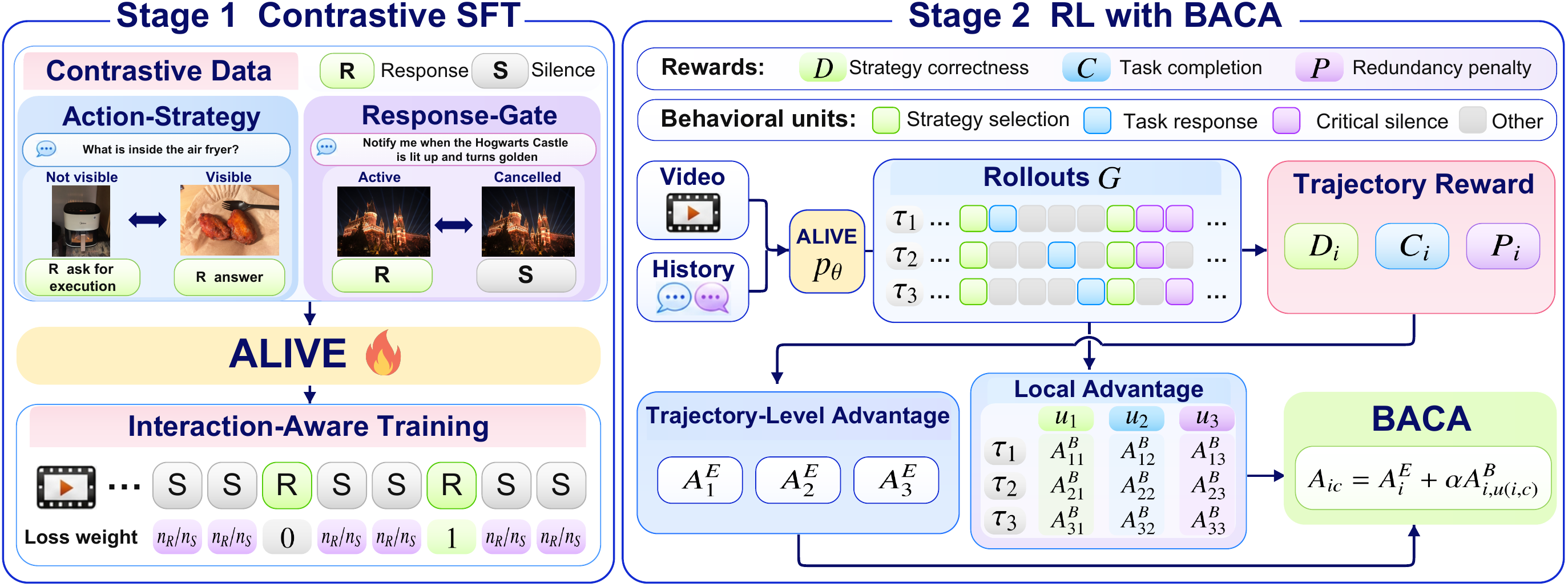}
    \caption{\textbf{Two-stage training of ALIVE.} \textbf{Left:} Contrastive SFT uses action-strategy and response-gate pairs to teach how available evidence and interaction history affect the appropriate output. $R$ and $S$ denote response and silence. Within each sampled window, the final response and all silence tokens receive supervision, while earlier responses are masked. \textbf{Right:} RL with Behavior-Aligned Credit Assignment (BACA) combines trajectory-level advantages with local advantages for strategy selection, task response, and critical silence. Trajectory rewards account for strategy correctness ($D$), task completion ($C$), and redundancy ($P$), and $\alpha$ weights the local advantage.}
    \label{fig:method}\vspace{-1.5em}
\end{figure*}

\noindent\textbf{Interaction-aware training.}
\label{sec:training}
We convert interaction trajectories into streaming training instances centered on annotated responses. For a response at time $a$, we sample a post-response offset $\delta$ uniformly from $\{0,\ldots,M\}$ and set the instance endpoint to $e=a+\delta$, capped before the next response or at the trajectory end. Extending instances beyond the response exposes the model to the transition back to silence. For response-gate contrasts, the designated comparison time takes precedence, ensuring that the instance reaches the critical response-or-silence decision.

We then retain a contiguous visual window ending at $e$. Its length $N$ is sampled uniformly between $N_{\min}$ and $\min(N_{\max},T_e)$, where $N_{\min}$ and $N_{\max}$ specify the minimum and maximum numbers of frames in the sampled visual window, and $T_e$ is the number of available frames up to $e$. Histories shorter than $N_{\min}$ are retained in full. Earlier user messages and substantive assistant responses remain as text-only context. Varying visual context while preserving interaction history helps the model adapt to sliding-window inference.

Within each instance, we supervise the final in-window response, including its control token and text, together with all silence tokens within the sampled visual window. Earlier responses serve as context and incur no loss. Each supervised output is conditioned only on the preceding observations and utterances.

To balance response and silence supervision, let $n_R$ and $n_S$ denote their respective numbers of supervised control tokens in a batch $\mathcal B$. Response control and text tokens receive unit weight, while silence tokens receive weight $n_R/n_S$. The SFT objective is
\begin{equation}
    \mathcal L_{\mathrm{SFT}}
    =
    -\frac{
    \sum_{j\in\mathcal B}m_jw_j
    \log p_\theta(y_j\mid x_{<j})
    }{
    \sum_{j\in\mathcal B}m_jw_j
    },
    \label{eq:objective}
\end{equation}
where $m_j$ selects supervised tokens, $w_j$ is the corresponding weight, and $x_{<j}$ denotes the preceding multimodal context.

\subsection{Reinforcement Learning with Behavior-Aligned Credit Assignment}
\label{sec:rl}

SFT establishes interaction behaviors from annotated trajectories. RL further refines these behaviors under histories containing the model's own outputs. In a multi-turn interaction, however, a trajectory may contain both successful and unsuccessful behaviors. Assigning the same trajectory-level advantage to every output provides limited guidance about which interaction requirements were satisfied. We therefore introduce \emph{Behavior-Aligned Credit Assignment} (BACA), which combines trajectory-level advantages with local advantages over annotated behavioral units.

\noindent\textbf{Streaming rollouts and trajectory reward.}
Starting from the SFT policy, we sample $G$ trajectories for each training episode. Video observations and user turns are replayed chronologically, and generated outputs become part of the subsequent interaction history. Policy sampling focuses on user turns and temporal regions around annotated response and critical-silence events. Outside these regions, observations continue to update the history while silence is simulated without contributing to the training loss. Annotations guide training-time scheduling and evaluation; the policy observes only the available video and interaction history.

Each sampled trajectory $\tau_i$ receives a reward
\begin{equation}
    R_i
    =
    D_iC_iP_i,
    \qquad
    P_i =
    \exp\!\left(
    -\lambda\frac{K_i}{\max(1,N_i^{\mathrm{req}})}
    \right),
    \label{eq:rl_reward}
\end{equation}
where $D_i$ evaluates the required interaction strategy and $C_i$ aggregates task completion, accounting for response correctness, timing, and compliance with the current task state, including annotated silence requirements. Semantic correctness is assessed by an LLM judge against the reference requirements. Temporal responses additionally receive credit according to their alignment with the annotated target time. The redundancy penalty factor $P_i$ downweights trajectories with redundant responses, with $K_i$ denoting their count and $N_i^{\mathrm{req}}$ the expected response count.

\noindent\textbf{Behavioral units.}
We construct local evaluation units from annotated user turns, response targets, and silence conditions. These units capture three complementary aspects of interaction: \emph{strategy selection}, which evaluates the appropriateness of the interaction strategy; \emph{task response}, which evaluates the correctness of answers or notifications and their timing when applicable; and \emph{critical silence}, which evaluates whether the model refrains from unwarranted responses during designated periods. For pre-event waiting, silence receives local credit only when the subsequent target also receives positive completion credit.

A behavioral unit may span one or several model calls, including both responses and silence. Its extent is determined by the relevant interaction requirement rather than a fixed number of dialogue turns. Corresponding units are aligned across trajectories by their annotation identities, allowing comparison even when sampled trajectories produce different numbers of responses. Each generated call is assigned to at most one local unit.

\begin{table*}[!t]
\centering
\caption{Results on public streaming video benchmarks. We report accuracy (\%) for OVO-Bench, StreamingBench, and IPIBench, and PAUC for ProactiveVideoQA. OVO-Bench: Real-Time Visual Perception (RTVP), Backward Tracing (BT), and Forward Active Responding (FAR). StreamingBench: Real-Time Visual Understanding (RTVU), Omni-Source Understanding (OSU), and Contextual Understanding (CU). IPIBench: Proactive Monitoring (PM), Proactive Task Management (PTM), and Interleaved Reactive-Proactive Requests (IRP). ProactiveVideoQA: Web-Video QA (WEB), Egocentric Video QA (EGO), TV-Series Video QA (TV), and Video Anomaly Detection (VAD). Avg denotes each task group's score, and All denotes each benchmark's overall score. Best model scores are bold.}
\label{tab:streaming_benchmarks}
\begingroup
\footnotesize
\renewcommand{\textbf}[1]{{\sffamily\bfseries #1}}
\setlength{\tabcolsep}{1.2pt}
\renewcommand{\arraystretch}{1.10}
\begin{tabular*}{\textwidth}{@{\extracolsep{\fill}}l*{16}{c}@{}}
\toprule
\multicolumn{17}{@{}l}{\textbf{OVO-Bench}~\citep{li2025ovobenchfarvideollmsrealworld}} \\
\midrule
\multirow{2}{*}{Model} & \multicolumn{7}{c}{RTVP} & \multicolumn{4}{c}{BT} & \multicolumn{4}{c}{FAR} & \multirow{2}{*}{All} \\
\cmidrule(lr){2-8}\cmidrule(lr){9-12}\cmidrule(lr){13-16}
 & OCR & ACR & ATR & STU & FPD & OJR & Avg & EPM & ASI & HLD & Avg & REC & SSR & CRR & Avg & \\
\midrule
VideoLLM-Online & 8.1 & 23.9 & 12.1 & 14.0 & 45.5 & 21.2 & 20.8 & 22.2 & 18.8 & 12.2 & 17.7 & -- & -- & -- & -- & -- \\
Flash-VStream & 24.2 & 29.4 & 28.5 & 33.7 & 25.7 & 28.8 & 28.4 & 39.1 & 37.2 & 5.9 & 27.4 & 8.0 & 67.3 & 60.0 & 45.1 & 33.6 \\
Dispider & 57.7 & 49.5 & 62.1 & 44.9 & 61.4 & 51.6 & 54.6 & 48.5 & 55.4 & 4.3 & 36.1 & 18.1 & 37.4 & 48.8 & 34.7 & 41.8 \\
VideoChat3 & 89.3 & 70.6 & 76.7 & 63.5 & 70.3 & 65.8 & 72.7 & 55.2 & 67.6 & 32.8 & 51.9 & 30.1 & 68.2 & 57.1 & 51.8 & 58.8 \\
JoyAI-VL-Interaction & 89.9 & 67.9 & 76.7 & 62.9 & 75.3 & 70.7 & 73.9 & 57.9 & 68.9 & 38.2 & 55.0 & 31.1 & 70.3 & 57.1 & 52.8 & 60.6 \\
AURA & 89.9 & 79.8 & 80.2 & \textbf{70.2} & \textbf{77.2} & \textbf{81.5} & \textbf{79.8} & 54.9 & 67.6 & 58.6 & 60.4 & 30.4 & \textbf{75.8} & 61.3 & \textbf{55.8} & 65.3 \\
Qwen3-VL-8B & 75.2 & 58.7 & 75.0 & 55.1 & 67.3 & 62.0 & 65.5 & 54.5 & \textbf{70.9} & 9.7 & 45.1 & \textbf{44.3} & 65.2 & 46.7 & 52.0 & 54.2 \\
\textbf{ALIVE} & \textbf{94.6} & \textbf{80.7} & \textbf{81.0} & 69.1 & 73.3 & 78.8 & 79.6 & \textbf{66.3} & 64.2 & \textbf{87.1} & \textbf{72.5} & 32.1 & 69.5 & \textbf{62.1} & 54.6 & \textbf{68.9} \\
\bottomrule
\end{tabular*}

\par\nointerlineskip\vspace{2pt}

\resizebox{\linewidth}{!}{%
\begin{tabular}{@{}l*{22}{c}@{}}
\multicolumn{23}{@{}l}{\textbf{StreamingBench}~\citep{lin2024streamingbenchassessinggapmllms}} \\
\midrule
\multirow{2}{*}{Model}
& \multicolumn{11}{c}{RTVU}
& \multicolumn{5}{c}{OSU}
& \multicolumn{5}{c}{CU}
& \multirow{2}{*}{All} \\
\cmidrule(lr){2-12}
\cmidrule(lr){13-17}
\cmidrule(lr){18-22}
& OP & CR & CS & ATP & EU & TR & PR & SU & ACP & CT & Avg
& ER & SCU & SD & MA & Avg
& MCU & ACU & SQA & PO & Avg & \\
\midrule
VideoLLM-Online & 39.1 & 40.1 & 34.5 & 31.1 & 46.0 & 32.4 & 31.5 & 34.2 & 42.5 & 27.9 & 36.0 & 31.2 & 26.5 & 24.1 & 32.0 & 28.5 & 29.2 & 24.2 & 30.8 & 3.9 & 26.6 & 32.5 \\
Flash-VStream & 25.9 & 43.6 & 24.9 & 23.9 & 27.3 & 13.1 & 18.5 & 25.2 & 23.9 & \textbf{48.7} & 23.2 & 25.9 & 24.9 & 25.6 & 28.4 & 26.0 & 25.2 & 24.8 & 26.8 & 2.0 & 24.1 & 24.0 \\
Dispider & 74.9 & 75.5 & 74.1 & 73.1 & 74.4 & 59.9 & 76.1 & 62.9 & 62.2 & 45.8 & 67.6 & 35.5 & 25.3 & 38.6 & 43.3 & 35.7 & 27.7 & 39.6 & 34.8 & 25.3 & 33.6 & 53.1 \\
VideoChat3 & 87.7 & 85.9 & 94.0 & 91.8 & 84.4 & 91.6 & 81.5 & \textbf{79.3} & 81.3 & 45.6 & 83.9 & 51.6 & 36.1 & 54.0 & 66.4 & 52.1 & 32.8 & 55.6 & 49.6 & 51.2 & 47.3 & 68.7 \\
JoyAI-VL-Interaction & 86.6 & 79.7 & 91.5 & 88.9 & \textbf{85.0} & 89.1 & 82.4 & 72.4 & 80.2 & 45.1 & 81.7 & 53.2 & 39.8 & 56.0 & 70.8 & 55.0 & 32.4 & 55.6 & 53.2 & 36.0 & 44.3 & 67.4 \\
AURA
& 87.5 & 84.4 & 93.4 & 89.5 & 83.2 & 87.9 & 86.1 & 76.4 & 82.4 & 47.7 & 83.2 & \textbf{56.4} & 48.4 & 62.8 & 80.4 & 62.0 & \textbf{54.8} & \textbf{70.8} & 57.2 & 53.2 & 59.0 & 73.1 \\
Qwen3-VL-8B & 89.1 & 78.1 & 95.6 & 90.5 & 81.9 & 90.0 & 86.1 & 72.0 & 75.1 & 45.1 & 82.0 & 50.0 & 34.5 & 49.6 & 62.0 & 49.0 & 35.2 & 53.2 & 51.6 & 46.4 & 46.6 & 66.8 \\
\textbf{ALIVE} & \textbf{90.5} & \textbf{87.5} & \textbf{96.9} & \textbf{93.8} & \textbf{85.0} & \textbf{95.0} & \textbf{88.0} & 78.9 & \textbf{86.4} & 46.6 & \textbf{86.5} & 54.8 & \textbf{50.2} & \textbf{74.8} & \textbf{81.6} & \textbf{65.4} & 48.0 & 70.0 & \textbf{60.0} & \textbf{63.6} & \textbf{60.4} & \textbf{76.0} \\
\bottomrule
\end{tabular}%
}

\par\nointerlineskip\vspace{2pt}

\begin{tabular*}{\textwidth}{@{\extracolsep{\fill}}l*{13}{c}@{}}
\multicolumn{14}{@{}l}{\textbf{IPIBench}~\citep{li2026ipibench}} \\
\midrule
\multirow{2}{*}{Model} & \multicolumn{4}{c}{PM} & \multicolumn{4}{c}{PTM} & \multicolumn{4}{c}{IRP} & \multirow{2}{*}{All} \\
\cmidrule(lr){2-5}\cmidrule(lr){6-9}\cmidrule(lr){10-13}
 & Timing & Under & Repeat & Avg & Cancel & Modify & Multi & Avg & R2P & RuP & RaP & Avg & \\
\midrule
VideoLLM-Online & 14.6 & 0.1 & 1.8 & 5.5 & 7.9 & 1.7 & 6.6 & 5.4 & 0.0 & 1.1 & 1.6 & 0.9 & 3.9 \\
Flash-VStream & 5.9 & 0.0 & 0.0 & 2.0 & 2.3 & 0.0 & 0.0 & 0.8 & 0.0 & 0.0 & 0.8 & 0.3 & 1.0 \\
Dispider & 18.7 & 0.7 & 1.8 & 7.1 & -- & -- & -- & -- & -- & -- & -- & -- & -- \\
VideoChat3 & 25.1 & 7.8 & 2.4 & 11.8 & 22.2 & 14.8 & 3.6 & 13.5 & 19.5 & 5.7 & 6.7 & 10.6 & 12.0 \\
JoyAI-VL-Interaction & 44.0 & 16.4 & 3.3 & 21.2 & 44.4 & 11.7 & 1.0 & 19.0 & 30.0 & 7.0 & 24.1 & 20.4 & 20.2 \\
AURA & 36.4 & 25.5 & 0.9 & 21.0 & 32.4 & 25.8 & 12.1 & 23.4 & 31.3 & 11.3 & 18.4 & 20.3 & 21.6 \\
Qwen3-VL-8B & 43.1 & 24.5 & 4.3 & 24.0 & 1.4 & 17.5 & 9.2 & 9.4 & 24.4 & 19.8 & 23.1 & 22.4 & 18.6 \\
\textbf{ALIVE} & \textbf{67.6} & \textbf{37.6} & \textbf{34.0} & \textbf{46.4} & \textbf{57.9} & \textbf{31.5} & \textbf{28.6} & \textbf{39.3} & \textbf{34.9} & \textbf{25.3} & \textbf{45.3} & \textbf{35.2} & \textbf{40.3} \\
\bottomrule
\end{tabular*}

\par\nointerlineskip\vspace{2pt}

\begin{tabular*}{\textwidth}{@{\extracolsep{\fill}}l*{5}{c}@{}}
\multicolumn{6}{@{}l}{\textbf{ProactiveVideoQA}~\citep{wang2025proactivevideoqacomprehensivebenchmarkevaluating}} \\
\midrule
Model & WEB & EGO & TV & VAD & All \\
\midrule
VideoLLM-Online & 25.9 & 25.0 & 18.3 & 25.0 & 23.6 \\
Flash-VStream & -- & -- & -- & -- & -- \\
Dispider & -- & -- & -- & -- & -- \\
VideoChat3 & 39.6 & 32.3 & 39.6 & \textbf{38.9} & 37.6 \\
JoyAI-VL-Interaction & 39.4 & 25.8 & 25.4 & 24.5 & 28.8 \\
AURA & 36.4 & 42.1 & 24.5 & 24.8 & 32.0 \\
Qwen3-VL-8B & 55.7 & 44.3 & 38.6 & 33.2 & 43.0 \\
\textbf{ALIVE} & \textbf{61.6} & \textbf{54.3} & \textbf{53.2} & 37.1 & \textbf{51.6} \\
\bottomrule
\end{tabular*}
\endgroup
\end{table*}

\noindent\textbf{Global and local advantages.}
We first compute a trajectory-level advantage by normalizing rewards within the group sampled for the same episode:
\begin{equation}
    A_i^{E}
    =
    \frac{R_i-\mu_R}{\sigma_R+\epsilon},
    \label{eq:grpo_advantage}
\end{equation}
where $\mu_R$ and $\sigma_R$ are the group mean and standard deviation. For behavioral unit $u$, let $s_{iu}$ denote its score in trajectory $i$, and let $\mathcal I_u$ contain the trajectories in which that unit is evaluated. Its local advantage is
\begin{equation}
    A_{iu}^{B}
    =
    \frac{s_{iu}-\mu_u}{\sigma_u+\epsilon},
    \qquad
    \mu_u
    =
    \frac{1}{|\mathcal I_u|}
    \sum_{i\in\mathcal I_u}s_{iu},
    \label{eq:behavior_advantage}
\end{equation}
where $\sigma_u$ is computed over the same group.

For model call $c$ in trajectory $i$, we combine the two signals as
\begin{equation}
    A_{ic}
    =
    A_i^{E}
    +
    \alpha A_{i,u(i,c)}^{B},
    \label{eq:combined_advantage}
\end{equation}
where $u(i,c)$ denotes its assigned behavioral unit and $\alpha$ weights the local advantage. Calls outside all units receive only the trajectory-level advantage. All generated tokens within a call share the resulting advantage. This preserves supervision on overall interaction quality while providing distinct learning signals for corresponding behavioral requirements.

\noindent\textbf{Policy optimization.}
We optimize the policy with the combined advantages:
\begin{equation}
    \mathcal L_{\mathrm{RL}}
    =
    -\frac{1}{B}
    \sum_{i=1}^{B}
    \frac{1}{n_i}
    \sum_{c=1}^{n_i}
    \frac{\operatorname{sg}(A_{ic})}{\ell_{ic}}
    \log p_\theta(o_{ic}\mid h_{ic})
    +
    \beta\mathcal L_{\mathrm{KL}},
    \label{eq:rl_objective}
\end{equation}
where $B$ is the number of retained trajectories, $n_i$ is the number of sampled calls in trajectory $i$, $\ell_{ic}$ is the generated token count of output $o_{ic}$, and $h_{ic}$ denotes the preceding multimodal context. The sequence log-probability sums over its generated tokens, and $\operatorname{sg}$ denotes stop-gradient. The KL term regularizes the policy toward the frozen SFT model using the same normalization.

\section{Experiments and Results}
\label{sec:experiments}

In this section, we present the implementation details, evaluation protocols, main results, and ablation study.

\subsection{Implementation Details}
\label{sec:implementation}

We initialize ALIVE from Qwen3-VL-8B-Instruct~\citep{bai2025qwen3vltechnicalreport} and apply the two-stage training procedure described in \S\ref{sec:method}. To preserve general question-answering ability and reinforce the decision to answer directly when sufficient evidence is available, we augment ALIVE-SFT-288K and ALIVE-RL-3K with 274K and 800 curated reactive QA examples, respectively. These examples are drawn from LLaVA-Video~\citep{zhang2024llava}, Molmo2~\citep{clark2026molmo2}, and Ego4D~\citep{grauman2022ego4d}.
In both stages, we freeze the vision encoder and connector and update only the language model. SFT runs for one epoch with a batch size of 512 and a learning rate of $1 \times 10^{-5}$.
Videos are sampled at 1 FPS. For SFT instance construction, we set $N_{\min}=10$, $N_{\max}=30$, and the maximum post-response offset $M=4$ positions. RL uses $G=4$ rollouts per episode and $\alpha=1$.

\subsection{Evaluation Protocols}
\label{sec:metrics}
\label{sec:eval_protocol}

\noindent\textbf{Public Benchmarks.}
We evaluate on OVO-Bench~\citep{li2025ovobenchfarvideollmsrealworld}, StreamingBench~\citep{lin2024streamingbenchassessinggapmllms}, IPIBench~\citep{li2026ipibench}, and ProactiveVideoQA~\citep{wang2025proactivevideoqacomprehensivebenchmarkevaluating} using their original task-specific protocols. We report accuracy (\%) for the first three benchmarks and PAUC for ProactiveVideoQA, with $\omega=0.5$. We report official results where available. Otherwise, we adapt each model to the official evaluation code while preserving its native inference settings as closely as possible. Full task-level results are reported in Table~\ref{tab:streaming_benchmarks}.

\noindent\textbf{FSI-Bench.}
To evaluate feedback-seeking interaction in streaming video models (\S\ref{sec:task_definition}), we introduce FSI-Bench, comprising 1{,}235 episodes across four tasks: Ask-for-Monitor (AFM, 154), Ask-for-Information (AFI, 446), Ask-for-Execution (AFE, 459), and Direct Answer (DA, 176). The last task tests immediate answering when sufficient evidence is available.

We use two protocols at query time $t_0$. The Explicit Decision Protocol (EDP) requires a choice among Answer, Wait, and Request, while the Free-form Decision Protocol (FDP) infers this choice from the model's natural response using a fixed LLM judge. Both use causal video and feedback streams with identical scoring rules.

Let $d,a\in\{0,0.5,1\}$ denote initial decision and answer correctness scores, respectively. Decision scoring is binary, except that choosing Wait on AFE receives partial credit ($d=0.5$) for deferring an unsupported answer without requesting the required action. An LLM judge assigns $a$ by comparing the response with reference answers across all four tasks. Decision (D), Timing (T), and Accuracy (A) are defined as
\begin{equation}
\mathrm{D}=d,\qquad
\mathrm{T}=d\,\mathbf{1}[t_r\in\mathcal{W}],\qquad
\mathrm{A}=\mathrm{T}\,a,
\label{eq:fsi_metrics}
\end{equation}
where $t_r$ is the time of the first non-silent output after $t_0$, or the output at $t_0$ for Direct Answer. The valid response window is
\begin{equation}
\mathcal{W}=\begin{cases}
[t^{\star}-1,t^{\star}+1], & \text{AFM or AFE},\\
\{t_1\}, & \text{AFI},\\
\{t_0\}, & \text{Direct Answer}.
\end{cases}
\label{eq:fsi_window}
\end{equation}
Here, $t^{\star}$ is the earliest annotated time when visual evidence clearly supports an answer, and $t_1$ is the clarification arrival time. The $\pm1$-step tolerance accounts for uncertainty in the annotated temporal boundary. Missing replies receive zero T and A.

\begin{table*}[!t]
\centering
\caption{FSI-Bench results (\%) under the Explicit Decision Protocol (EDP) and Free-form Decision Protocol (FDP). AFM, AFI, AFE, and DA denote Ask-for-Monitor, Ask-for-Information, Ask-for-Execution, and Direct Answer, respectively. Each task reports Decision (D), Timing (T), and Accuracy (A). All is the four-task macro average. Best model scores are bold.}
\label{tab:fsibench}
\begingroup
\footnotesize
\renewcommand{\textbf}[1]{{\sffamily\bfseries #1}}
\setlength{\tabcolsep}{1.2pt}
\renewcommand{\arraystretch}{1.10}
\begin{tabular*}{\textwidth}{@{\extracolsep{\fill}}l*{15}{c}@{}}
\toprule
\multicolumn{16}{@{}l}{\textbf{EDP}} \\
\midrule
\multirow{2}{*}{Model} & \multicolumn{3}{c}{AFM} & \multicolumn{3}{c}{AFI} & \multicolumn{3}{c}{AFE} & \multicolumn{3}{c}{DA} & \multicolumn{3}{c}{All} \\
\cmidrule(lr){2-4}\cmidrule(lr){5-7}\cmidrule(lr){8-10}\cmidrule(lr){11-13}\cmidrule(lr){14-16}
 & D & T & A & D & T & A & D & T & A & D & T & A & D & T & A \\
\midrule
Human Level & --- & --- & --- & --- & --- & --- & --- & --- & --- & --- & --- & --- & --- & --- & --- \\
\midrule
\multicolumn{16}{@{}l}{\textit{Proprietary Models---Offline}} \\
GPT-6 Astra & 89.0 & 52.6 & 36.0 & 19.1 & 18.4 & 15.8 & 73.3 & 55.3 & \textbf{37.4} & 93.8 & 93.8 & \textbf{79.5} & 68.8 & 55.0 & 42.2 \\
GPT-5.6 Sol & 89.6 & 41.6 & 28.2 & 15.9 & 15.7 & 11.2 & 47.3 & 31.6 & 18.8 & 96.6 & 96.6 & 78.7 & 62.3 & 46.4 & 34.2 \\
Gemini 3.5 Flash & 87.0 & 20.1 & 13.0 & 0.9 & 0.9 & 0.9 & 32.0 & 19.9 & 12.4 & 97.2 & 97.2 & 77.3 & 54.3 & 34.5 & 25.9 \\
\midrule
\multicolumn{16}{@{}l}{\textit{Open-source Models---Offline}} \\
Qwen3.5-9B & 95.5 & 0.6 & 0.0 & 0.0 & 0.0 & 0.0 & 31.3 & 10.0 & 1.6 & 93.2 & 93.2 & 67.3 & 55.0 & 26.0 & 17.2 \\
Qwen3-VL-8B & 98.7 & 24.0 & 6.8 & 3.4 & 3.4 & 2.6 & 47.2 & 21.9 & 6.7 & 60.8 & 60.8 & 47.2 & 52.5 & 27.5 & 15.8 \\
InternVL3-8B & 98.7 & 4.5 & 2.3 & 0.0 & 0.0 & 0.0 & 46.2 & 19.9 & 5.9 & 67.6 & 67.6 & 50.0 & 53.1 & 23.0 & 14.5 \\
\midrule
\multicolumn{16}{@{}l}{\textit{Open-source Models---Online}} \\
VideoChat3 & \textbf{100.0} & 31.2 & 12.3 & 0.0 & 0.0 & 0.0 & 49.7 & 26.1 & 10.8 & 1.7 & 1.7 & 0.6 & 37.8 & 14.8 & 5.9 \\
JoyAI-VL-Interaction & 66.2 & 0.6 & 0.6 & 0.0 & 0.0 & 0.0 & 16.4 & 0.1 & 0.0 & 97.2 & 97.2 & 69.0 & 45.0 & 24.5 & 17.4 \\
AURA & \textbf{100.0} & 12.3 & 8.8 & 0.4 & 0.4 & 0.2 & 49.7 & 4.8 & 2.8 & 23.9 & 23.9 & 21.3 & 43.5 & 10.4 & 8.3 \\
\textbf{ALIVE} & \textbf{100.0} & \textbf{68.2} & \textbf{38.0} & \textbf{96.6} & \textbf{96.6} & \textbf{64.0} & \textbf{75.2} & \textbf{59.4} & 33.8 & \textbf{98.3} & \textbf{98.3} & 75.3 & \textbf{92.5} & \textbf{80.6} & \textbf{52.8} \\
\bottomrule
\end{tabular*}

\par\nointerlineskip\vspace{2pt}
\begin{tabular*}{\textwidth}{@{\extracolsep{\fill}}l*{15}{c}@{}}
\multicolumn{16}{@{}l}{\textbf{FDP}} \\
\midrule
\multirow{2}{*}{Model} & \multicolumn{3}{c}{AFM} & \multicolumn{3}{c}{AFI} & \multicolumn{3}{c}{AFE} & \multicolumn{3}{c}{DA} & \multicolumn{3}{c}{All} \\
\cmidrule(lr){2-4}\cmidrule(lr){5-7}\cmidrule(lr){8-10}\cmidrule(lr){11-13}\cmidrule(lr){14-16}
 & D & T & A & D & T & A & D & T & A & D & T & A & D & T & A \\
\midrule
Human Level & 98.0 & 96.0 & 88.0 & 100.0 & 100.0 & 90.0 & 100.0 & 94.0 & 86.0 & 100.0 & 100.0 & 90.0 & 99.5 & 97.5 & 88.5 \\
\midrule
\multicolumn{16}{@{}l}{\textit{Proprietary Models---Offline}} \\
GPT-6 Astra & 57.8 & 41.6 & 28.2 & 0.0 & 0.0 & 0.0 & 18.7 & 15.8 & 10.5 & \textbf{100.0} & \textbf{100.0} & 81.8 & 44.1 & 39.3 & 30.1 \\
GPT-5.6 Sol & 44.8 & 29.2 & 19.2 & 0.0 & 0.0 & 0.0 & 14.5 & 9.8 & 6.0 & \textbf{100.0} & \textbf{100.0} & 79.8 & 39.8 & 34.8 & 26.3 \\
Gemini 3.5 Flash & 28.6 & 6.5 & 4.9 & 0.0 & 0.0 & 0.0 & 12.0 & 8.4 & 5.2 & \textbf{100.0} & \textbf{100.0} & \textbf{82.7} & 35.1 & 28.7 & 23.2 \\
\midrule
\multicolumn{16}{@{}l}{\textit{Open-source Models---Offline}} \\
Qwen3.5-9B & 36.4 & 0.6 & 0.0 & 0.0 & 0.0 & 0.0 & 10.5 & 4.2 & 0.7 & \textbf{100.0} & \textbf{100.0} & 73.6 & 36.7 & 26.2 & 18.6 \\
Qwen3-VL-8B & 22.1 & 7.1 & 2.3 & 0.0 & 0.0 & 0.0 & 11.7 & 6.2 & 2.2 & 97.7 & 97.7 & 67.9 & 32.9 & 27.8 & 18.1 \\
InternVL3-8B & 0.6 & 0.0 & 0.0 & 0.0 & 0.0 & 0.0 & 1.2 & 0.2 & 0.0 & \textbf{100.0} & \textbf{100.0} & 65.1 & 25.5 & 25.1 & 16.3 \\
\midrule
\multicolumn{16}{@{}l}{\textit{Open-source Models---Online}} \\
VideoChat3 & \textbf{99.4} & 24.0 & 11.7 & 0.0 & 0.0 & 0.0 & 46.9 & 19.2 & 6.6 & 13.1 & 13.1 & 6.2 & 39.8 & 14.1 & 6.1 \\
JoyAI-VL-Interaction & 0.6 & 0.0 & 0.0 & 0.0 & 0.0 & 0.0 & 0.5 & 0.0 & 0.0 & \textbf{100.0} & \textbf{100.0} & 71.9 & 25.3 & 25.0 & 18.0 \\
AURA & 0.6 & 0.0 & 0.0 & 0.0 & 0.0 & 0.0 & 0.5 & 0.1 & 0.1 & 99.4 & 99.4 & 72.4 & 25.2 & 24.9 & 18.1 \\
\textbf{ALIVE} & 92.2 & \textbf{62.3} & \textbf{36.7} & \textbf{97.5} & \textbf{97.5} & \textbf{64.9} & \textbf{81.9} & \textbf{66.1} & \textbf{38.7} & \textbf{100.0} & \textbf{100.0} & 75.6 & \textbf{92.9} & \textbf{81.5} & \textbf{54.0} \\
\bottomrule
\end{tabular*}

\endgroup
\end{table*}

\subsection{Main Results}
\label{sec:main_results}
\noindent\textbf{Public Benchmarks.}
We compare ALIVE with VideoLLM-Online~\citep{chen2024videollmonlineonlinevideolarge}, Flash-VStream~\citep{zhang2025flash}, Dispider~\citep{qian2025dispiderenablingvideollms}, VideoChat3~\citep{li2026videochat3}, JoyAI-VL-Interaction~\citep{yao2026joyaivlinteractionrealtimevisionlanguageinteraction}, AURA~\citep{lu2026auraalwaysonunderstandingrealtime}, and Qwen3-VL-8B~\citep{bai2025qwen3vltechnicalreport} on four public benchmarks (Table~\ref{tab:streaming_benchmarks}). ALIVE achieves state-of-the-art overall performance among the compared models on all four benchmarks. The largest gain is on IPIBench, where ALIVE scores 40.3, surpassing the strongest baseline, AURA, by 18.7 points. These results demonstrate that ALIVE combines strong streaming video understanding with enhanced interactive capabilities, including managing concurrent user tasks and adapting to changes in their states.

\noindent\textbf{FSI-Bench.}
We evaluate proprietary models (GPT-6 Astra~\citep{openai2026gpt6astra}, GPT-5.6 Sol~\citep{openai2026gpt56}, and Gemini 3.5 Flash~\citep{googledeepmind2026gemini35flash}), open-source models (Qwen3.5-9B~\citep{qwen3.5}, Qwen3-VL-8B~\citep{bai2025qwen3vltechnicalreport}, and InternVL3-8B~\citep{zhu2025internvl3}), and streaming models (VideoChat3~\citep{li2026videochat3}, JoyAI-VL-Interaction~\citep{yao2026joyaivlinteractionrealtimevisionlanguageinteraction}, and AURA~\citep{lu2026auraalwaysonunderstandingrealtime}) alongside ALIVE on our FSI-Bench (Table~\ref{tab:fsibench}). ALIVE achieves overall Accuracy scores of 52.8 under EDP and 54.0 under FDP, exceeding GPT-6 Astra, the strongest model baseline, by 10.6 and 23.9 points respectively. Proprietary models perform well on DA but struggle on AFI and AFE. Apart from ALIVE, the evaluated open-source streaming models perform poorly across all three feedback-seeking tasks. ALIVE's gains reflect progress from passive video understanding toward streaming interaction that actively seeks user input to resolve missing information. We evaluate human performance under FDP as a reference for natural interaction. Despite its leading performance, ALIVE remains far below the human score of 88.5, leaving substantial room for improvement.

\subsection{Ablation Study}
\label{sec:ablation}

\begin{table}[t]
\centering
\caption{Ablation of contrastive SFT and BACA. We report overall scores on OVO-Bench (OVO), StreamingBench (SB), IPIBench (IPI), and ProactiveVideoQA (PVQA), and four-task Accuracy on FSI-Bench under EDP and FDP. Best scores are bold.}
\label{tab:ablation}
\begingroup
\footnotesize
\setlength{\tabcolsep}{2pt}
\renewcommand{\arraystretch}{1.08}
\begin{tabular*}{\columnwidth}{@{\extracolsep{\fill}}l*{6}{c}@{}}
\toprule
\multirow{2}{*}{Configuration} & \multirow{2}{*}{OVO} & \multirow{2}{*}{SB} & \multirow{2}{*}{IPI} & \multirow{2}{*}{PVQA} & \multicolumn{2}{c}{FSI-Bench} \\
\cmidrule(lr){6-7}
 & & & & & EDP & FDP \\
\midrule
SFT w/o contrasts & 65.8 & 73.1 & 37.5 & 47.1 & 48.1 & 39.4 \\
Contrastive SFT & 68.4 & 74.9 & 38.8 & 48.6 & 51.1 & 49.3 \\
RL w/o BACA & 68.7 & 75.7 & 38.2 & 50.4 & 51.7 & 51.5 \\
\textbf{ALIVE} & \textbf{68.9} & \textbf{76.0} & \textbf{40.3} & \textbf{51.6} & \textbf{52.8} & \textbf{54.0} \\
\bottomrule
\end{tabular*}
\endgroup
\end{table}

Table~\ref{tab:ablation} compares four configurations: SFT without contrastive examples, contrastive SFT, trajectory-only RL, and the full ALIVE model. Both RL variants start from contrastive SFT and use the reward in Eq.~\ref{eq:rl_reward}. RL w/o BACA uses GRPO-style trajectory-level advantages alone, equivalent to setting $\alpha=0$ in Eq.~\ref{eq:combined_advantage}. ALIVE additionally incorporates local advantages over behavioral units.

Contrastive SFT improves all six aggregate scores, with the largest gain on FSI-Bench under FDP (+9.9 points). ALIVE achieves the highest aggregate scores across all benchmarks and both FSI-Bench protocols. These results show that contrastive SFT and BACA jointly improve ALIVE's streaming interaction capabilities.

\FloatBarrier

\section{Conclusion and Limitations}
\label{sec:conclusion}

We formulate interactive streaming video understanding around proactive monitoring, streaming task coordination, and feedback-seeking interaction, and introduce ALIVE to support these capabilities. We train the model on ALIVE-SFT-288K and ALIVE-RL-3K through contrastive supervised fine-tuning and reinforcement learning with Behavior-Aligned Credit Assignment (BACA). ALIVE achieves state-of-the-art overall performance among the compared models on four public benchmarks and FSI-Bench, demonstrating strong capabilities in both streaming video understanding and interaction. We believe ALIVE provides an empirical foundation for developing multimodal assistants that continuously perceive the world and sustain meaningful interactions with users as their needs evolve. 

ALIVE remains a vision-language model, and enabling spoken interaction requires external automatic speech recognition (ASR) and text-to-speech (TTS) components. Extending this framework to native omni-modal models is a promising direction toward more natural streaming interaction across vision, language, and audio.

\par
\clearpage
\bibliographystyle{plainnat}
\bibliography{main}

@misc{bai2025qwen3vltechnicalreport,
      title={Qwen3-VL Technical Report}, 
      author={Shuai Bai and Yuxuan Cai and Ruizhe Chen and Keqin Chen and Xionghui Chen and Zesen Cheng and Lianghao Deng and Wei Ding and Chang Gao and Chunjiang Ge and Wenbin Ge and Zhifang Guo and Qidong Huang and Jie Huang and Fei Huang and Binyuan Hui and Shutong Jiang and Zhaohai Li and Mingsheng Li and Mei Li and Kaixin Li and Zicheng Lin and Junyang Lin and Xuejing Liu and Jiawei Liu and Chenglong Liu and Yang Liu and Dayiheng Liu and Shixuan Liu and Dunjie Lu and Ruilin Luo and Chenxu Lv and Rui Men and Lingchen Meng and Xuancheng Ren and Xingzhang Ren and Sibo Song and Yuchong Sun and Jun Tang and Jianhong Tu and Jianqiang Wan and Peng Wang and Pengfei Wang and Qiuyue Wang and Yuxuan Wang and Tianbao Xie and Yiheng Xu and Haiyang Xu and Jin Xu and Zhibo Yang and Mingkun Yang and Jianxin Yang and An Yang and Bowen Yu and Fei Zhang and Hang Zhang and Xi Zhang and Bo Zheng and Humen Zhong and Jingren Zhou and Fan Zhou and Jing Zhou and Yuanzhi Zhu and Ke Zhu},
      year={2025},
      eprint={2511.21631},
      archivePrefix={arXiv},
      primaryClass={cs.CV},
      url={https://arxiv.org/abs/2511.21631}, 
}

@misc{chen2024videollmonlineonlinevideolarge,
      title={VideoLLM-online: Online Video Large Language Model for Streaming Video}, 
      author={Joya Chen and Zhaoyang Lv and Shiwei Wu and Kevin Qinghong Lin and Chenan Song and Difei Gao and Jia-Wei Liu and Ziteng Gao and Dongxing Mao and Mike Zheng Shou},
      year={2024},
      eprint={2406.11816},
      archivePrefix={arXiv},
      primaryClass={cs.CV},
      url={https://arxiv.org/abs/2406.11816}, 
}

@misc{qian2025dispiderenablingvideollms,
      title={Dispider: Enabling Video LLMs with Active Real-Time Interaction via Disentangled Perception, Decision, and Reaction}, 
      author={Rui Qian and Shuangrui Ding and Xiaoyi Dong and Pan Zhang and Yuhang Zang and Yuhang Cao and Dahua Lin and Jiaqi Wang},
      year={2025},
      eprint={2501.03218},
      archivePrefix={arXiv},
      primaryClass={cs.CV},
      url={https://arxiv.org/abs/2501.03218}, 
}

@misc{wang2025mmduet2enhancingproactiveinteraction,
      title={MMDuet2: Enhancing Proactive Interaction of Video MLLMs with Multi-Turn Reinforcement Learning}, 
      author={Yueqian Wang and Songxiang Liu and Disong Wang and Nuo Xu and Guanglu Wan and Huishuai Zhang and Dongyan Zhao},
      year={2025},
      eprint={2512.06810},
      archivePrefix={arXiv},
      primaryClass={cs.CV},
      url={https://arxiv.org/abs/2512.06810}, 
}

@misc{xia2026streamingvideoinstructiontuning,
      title={Streaming Video Instruction Tuning}, 
      author={Jiaer Xia and Peixian Chen and Mengdan Zhang and Xing Sun and Kaiyang Zhou},
      year={2026},
      eprint={2512.21334},
      archivePrefix={arXiv},
      primaryClass={cs.CV},
      url={https://arxiv.org/abs/2512.21334}, 
}

@misc{lu2026auraalwaysonunderstandingrealtime,
      title={AURA: Always-On Understanding and Real-Time Assistance via Video Streams}, 
      author={Xudong Lu and Yang Bo and Jinpeng Chen and Shuhan Li and Xintong Guo and Huankang Guan and Fang Liu and Dunyuan Xu and Peiwen Sun and Heyang Sun and Rui Liu and Hongsheng Li},
      year={2026},
      eprint={2604.04184},
      archivePrefix={arXiv},
      primaryClass={cs.CV},
      url={https://arxiv.org/abs/2604.04184}, 
}

@misc{yao2026joyaivlinteractionrealtimevisionlanguageinteraction,
      title={JoyAI-VL-Interaction: Real-Time Vision-Language Interaction Intelligence}, 
      author={Dingyu Yao and Junhao Zhou and Chenxu Yang and Chuanyu Qin and Haowen Hou and Zheming Liang and Congcong Wang and Yuhang Cao and Shenglong Ye and Shuai Xie and Shuhuan Gu and Haoyang Huang and Qingyi Si and Nan Duan and Jiaqi Wang},
      year={2026},
      eprint={2606.14777},
      archivePrefix={arXiv},
      primaryClass={cs.CV},
      url={https://arxiv.org/abs/2606.14777}, 
}

@misc{wang2025streambridgeturningofflinevideo,
      title={StreamBridge: Turning Your Offline Video Large Language Model into a Proactive Streaming Assistant}, 
      author={Haibo Wang and Bo Feng and Zhengfeng Lai and Mingze Xu and Shiyu Li and Weifeng Ge and Afshin Dehghan and Meng Cao and Ping Huang},
      year={2025},
      eprint={2505.05467},
      archivePrefix={arXiv},
      primaryClass={cs.CV},
      url={https://arxiv.org/abs/2505.05467}, 
}

@misc{xu2026streamingvlmrealtimeunderstandinginfinite,
      title={StreamingVLM: Real-Time Understanding for Infinite Video Streams}, 
      author={Ruyi Xu and Guangxuan Xiao and Yukang Chen and Liuning He and Yao Lu and Song Han},
      year={2026},
      eprint={2510.09608},
      archivePrefix={arXiv},
      primaryClass={cs.CV},
      url={https://arxiv.org/abs/2510.09608}, 
}

@misc{lin2024streamingbenchassessinggapmllms,
      title={StreamingBench: Assessing the Gap for MLLMs to Achieve Streaming Video Understanding}, 
      author={Junming Lin and Zheng Fang and Chi Chen and Zihao Wan and Fuwen Luo and Peng Li and Yang Liu and Maosong Sun},
      year={2024},
      eprint={2411.03628},
      archivePrefix={arXiv},
      primaryClass={cs.CV},
      url={https://arxiv.org/abs/2411.03628}, 
}

@misc{li2025ovobenchfarvideollmsrealworld,
      title={OVO-Bench: How Far is Your Video-LLMs from Real-World Online Video Understanding?}, 
      author={Yifei Li and Junbo Niu and Ziyang Miao and Chunjiang Ge and Yuanhang Zhou and Qihao He and Xiaoyi Dong and Haodong Duan and Shuangrui Ding and Rui Qian and Pan Zhang and Yuhang Zang and Yuhang Cao and Conghui He and Jiaqi Wang},
      year={2025},
      eprint={2501.05510},
      archivePrefix={arXiv},
      primaryClass={cs.CV},
      url={https://arxiv.org/abs/2501.05510}, 
}

@misc{xiong2025streamingvideounderstandingmultiround,
      title={Streaming Video Understanding and Multi-round Interaction with Memory-enhanced Knowledge}, 
      author={Haomiao Xiong and Zongxin Yang and Jiazuo Yu and Yunzhi Zhuge and Lu Zhang and Jiawen Zhu and Huchuan Lu},
      year={2025},
      eprint={2501.13468},
      archivePrefix={arXiv},
      primaryClass={cs.CV},
      url={https://arxiv.org/abs/2501.13468}, 
}

@misc{wang2025proactivevideoqacomprehensivebenchmarkevaluating,
      title={ProactiveVideoQA: A Comprehensive Benchmark Evaluating Proactive Interactions in Video Large Language Models}, 
      author={Yueqian Wang and Xiaojun Meng and Yifan Wang and Huishuai Zhang and Dongyan Zhao},
      year={2025},
      eprint={2507.09313},
      archivePrefix={arXiv},
      primaryClass={cs.CV},
      url={https://arxiv.org/abs/2507.09313}, 
}

@article{li2026ipibench,
  title={IPIBench: Evaluating Interactive Proactive Intelligence of MLLMs under Continuous Streams},
  author={Li, Jinzhao and Chen, Yinuo and Song, Wenxuan and Lei, Yijia and Zhang, Yichi and Yan, Honglei and Pan, Panwang and Liu, Miao},
  journal={arXiv preprint arXiv:2605.27074},
  year={2026}
}

@inproceedings{zhang2025proactive,
  title     = {Proactive Assistant Dialogue Generation from Streaming Egocentric Videos},
  author    = {Zhang, Yichi and Dong, Xin Luna and Lin, Zhaojiang and Madotto, Andrea
               and Kumar, Anuj and Damavandi, Babak and Chai, Joyce and Moon, Seungwhan},
  booktitle = {Proceedings of the 2025 Conference on Empirical Methods in Natural Language Processing},
  pages     = {12044--12068},
  year      = {2025},
  publisher = {Association for Computational Linguistics},
  doi       = {10.18653/v1/2025.emnlp-main.605},
  url       = {https://aclanthology.org/2025.emnlp-main.605/}
}

@article{li2026videochat3,
  title={VideoChat3: Fully Open Video MLLM for Efficient and Generalist Video Understanding},
  author={Li, Xinhao and Zhu, Yuhan and Zeng, Xiangyu and Dong, Yuhao and Wu, Haoning and Zhang, Zhiqiu and Yang, Yuandong and Ma, Changlian and Zhang, Qingyu and Shi, Yansong and others},
  journal={arXiv preprint arXiv:2607.14935},
  year={2026}
}

@misc{ye2025realwebassistbenchmarklonghorizonweb,
      title={RealWebAssist: A Benchmark for Long-Horizon Web Assistance with Real-World Users}, 
      author={Suyu Ye and Haojun Shi and Darren Shih and Hyokun Yun and Tanya Roosta and Tianmin Shu},
      year={2025},
      eprint={2504.10445},
      archivePrefix={arXiv},
      primaryClass={cs.AI},
      url={https://arxiv.org/abs/2504.10445}, 
}

@misc{lu2024proactiveagentshiftingllm,
      title={Proactive Agent: Shifting LLM Agents from Reactive Responses to Active Assistance}, 
      author={Yaxi Lu and Shenzhi Yang and Cheng Qian and Guirong Chen and Qinyu Luo and Yesai Wu and Huadong Wang and Xin Cong and Zhong Zhang and Yankai Lin and Weiwen Liu and Yasheng Wang and Zhiyuan Liu and Fangming Liu and Maosong Sun},
      year={2024},
      eprint={2410.12361},
      archivePrefix={arXiv},
      primaryClass={cs.AI},
      url={https://arxiv.org/abs/2410.12361}, 
}

@inproceedings{grauman2022ego4d,
  title={Ego4d: Around the world in 3,000 hours of egocentric video},
  author={Grauman, Kristen and Westbury, Andrew and Byrne, Eugene and Chavis, Zachary and Furnari, Antonino and Girdhar, Rohit and Hamburger, Jackson and Jiang, Hao and Liu, Miao and Liu, Xingyu and others},
  booktitle={Proceedings of the IEEE/CVF conference on computer vision and pattern recognition},
  pages={18995--19012},
  year={2022}
}

@article{zhang2024llava,
  title={Llava-video: Video instruction tuning with synthetic data},
  author={Zhang, Yuanhan and Wu, Jinming and Li, Wei and Li, Bo and Ma, Zejun and Liu, Ziwei and Li, Chunyuan},
  journal={arXiv preprint arXiv:2410.02713},
  year={2024}
}

@article{wu2024star,
  title={Star: A benchmark for situated reasoning in real-world videos},
  author={Wu, Bo and Yu, Shoubin and Chen, Zhenfang and Tenenbaum, Joshua B and Gan, Chuang},
  journal={arXiv preprint arXiv:2405.09711},
  year={2024}
}

@misc{zhu2026timelens2generalistvideotemporal,
      title={TimeLens2: Generalist Video Temporal Grounding with Multimodal LLMs}, 
      author={Yuhan Zhu and Changlian Ma and Xiangyu Zeng and Xinhao Li and Zhiqiu Zhang and Songze Li and Jun Zhang and Tianxiang Jiang and Yuandong Yang and Ziang Yan and Zikang Wang and Xinyu Chen and Haoran Chen and Shaowei Zhang and Limin Wang},
      year={2026},
      eprint={2607.17423},
      archivePrefix={arXiv},
      primaryClass={cs.CV},
      url={https://arxiv.org/abs/2607.17423}, 
}

@misc{zhang2026timelensrethinkingvideotemporal,
      title={TimeLens: Rethinking Video Temporal Grounding with Multimodal LLMs}, 
      author={Jun Zhang and Teng Wang and Yuying Ge and Yixiao Ge and Xinhao Li and Ying Shan and Limin Wang},
      year={2026},
      eprint={2512.14698},
      archivePrefix={arXiv},
      primaryClass={cs.CV},
      url={https://arxiv.org/abs/2512.14698}, 
}

@misc{ma2024nymeriamassivecollectionmultimodal,
      title={Nymeria: A Massive Collection of Multimodal Egocentric Daily Motion in the Wild}, 
      author={Lingni Ma and Yuting Ye and Fangzhou Hong and Vladimir Guzov and Yifeng Jiang and Rowan Postyeni and Luis Pesqueira and Alexander Gamino and Vijay Baiyya and Hyo Jin Kim and Kevin Bailey and David Soriano Fosas and C. Karen Liu and Ziwei Liu and Jakob Engel and Renzo De Nardi and Richard Newcombe},
      year={2024},
      eprint={2406.09905},
      archivePrefix={arXiv},
      primaryClass={cs.CV},
      url={https://arxiv.org/abs/2406.09905}, 
}

@misc{ge2026densestep2mscalabletrainingfreepipeline,
      title={DenseStep2M: A Scalable, Training-Free Pipeline for Dense Instructional Video Annotation}, 
      author={Mingji Ge and Qirui Chen and Zeqian Li and Weidi Xie},
      year={2026},
      eprint={2604.26565},
      archivePrefix={arXiv},
      primaryClass={cs.CV},
      url={https://arxiv.org/abs/2604.26565}, 
}

@misc{yuan2025videorefersuiteadvancingspatialtemporal,
      title={VideoRefer Suite: Advancing Spatial-Temporal Object Understanding with Video LLM}, 
      author={Yuqian Yuan and Hang Zhang and Wentong Li and Zesen Cheng and Boqiang Zhang and Long Li and Xin Li and Deli Zhao and Wenqiao Zhang and Yueting Zhuang and Jianke Zhu and Lidong Bing},
      year={2025},
      eprint={2501.00599},
      archivePrefix={arXiv},
      primaryClass={cs.CV},
      url={https://arxiv.org/abs/2501.00599}, 
}

@misc{maaz2024videogptintegratingimagevideo,
      title={VideoGPT+: Integrating Image and Video Encoders for Enhanced Video Understanding}, 
      author={Muhammad Maaz and Hanoona Rasheed and Salman Khan and Fahad Khan},
      year={2024},
      eprint={2406.09418},
      archivePrefix={arXiv},
      primaryClass={cs.CV},
      url={https://arxiv.org/abs/2406.09418}, 
}

@misc{tang2019coinlargescaledatasetcomprehensive,
      title={COIN: A Large-scale Dataset for Comprehensive Instructional Video Analysis}, 
      author={Yansong Tang and Dajun Ding and Yongming Rao and Yu Zheng and Danyang Zhang and Lili Zhao and Jiwen Lu and Jie Zhou},
      year={2019},
      eprint={1903.02874},
      archivePrefix={arXiv},
      primaryClass={cs.CV},
      url={https://arxiv.org/abs/1903.02874}, 
}

@misc{damen2020epickitchensdatasetcollectionchallenges,
      title={The EPIC-KITCHENS Dataset: Collection, Challenges and Baselines}, 
      author={Dima Damen and Hazel Doughty and Giovanni Maria Farinella and Sanja Fidler and Antonino Furnari and Evangelos Kazakos and Davide Moltisanti and Jonathan Munro and Toby Perrett and Will Price and Michael Wray},
      year={2020},
      eprint={2005.00343},
      archivePrefix={arXiv},
      primaryClass={cs.CV},
      url={https://arxiv.org/abs/2005.00343}, 
}

@misc{yang2024vriptvideoworththousands,
      title={Vript: A Video Is Worth Thousands of Words}, 
      author={Dongjie Yang and Suyuan Huang and Chengqiang Lu and Xiaodong Han and Haoxin Zhang and Yan Gao and Yao Hu and Hai Zhao},
      year={2024},
      eprint={2406.06040},
      archivePrefix={arXiv},
      primaryClass={cs.CV},
      url={https://arxiv.org/abs/2406.06040}, 
}

@misc{zhao2019hacshumanactionclips,
      title={HACS: Human Action Clips and Segments Dataset for Recognition and Temporal Localization}, 
      author={Hang Zhao and Antonio Torralba and Lorenzo Torresani and Zhicheng Yan},
      year={2019},
      eprint={1712.09374},
      archivePrefix={arXiv},
      primaryClass={cs.CV},
      url={https://arxiv.org/abs/1712.09374}, 
}

@misc{lei2021qvhighlightsdetectingmomentshighlights,
      title={QVHighlights: Detecting Moments and Highlights in Videos via Natural Language Queries}, 
      author={Jie Lei and Tamara L. Berg and Mohit Bansal},
      year={2021},
      eprint={2107.09609},
      archivePrefix={arXiv},
      primaryClass={cs.CV},
      url={https://arxiv.org/abs/2107.09609}, 
}

@misc{han2025shot2storynewbenchmarkcomprehensive,
      title={Shot2Story: A New Benchmark for Comprehensive Understanding of Multi-shot Videos}, 
      author={Mingfei Han and Linjie Yang and Xiaojun Chang and Lina Yao and Heng Wang},
      year={2025},
      eprint={2312.10300},
      archivePrefix={arXiv},
      primaryClass={cs.CV},
      url={https://arxiv.org/abs/2312.10300}, 
}

@misc{sigurdsson2018charadesegolargescaledatasetpaired,
      title={Charades-Ego: A Large-Scale Dataset of Paired Third and First Person Videos}, 
      author={Gunnar A. Sigurdsson and Abhinav Gupta and Cordelia Schmid and Ali Farhadi and Karteek Alahari},
      year={2018},
      eprint={1804.09626},
      archivePrefix={arXiv},
      primaryClass={cs.CV},
      url={https://arxiv.org/abs/1804.09626}, 
}

@misc{bain2022frozentimejointvideo,
      title={Frozen in Time: A Joint Video and Image Encoder for End-to-End Retrieval}, 
      author={Max Bain and Arsha Nagrani and Gül Varol and Andrew Zisserman},
      year={2022},
      eprint={2104.00650},
      archivePrefix={arXiv},
      primaryClass={cs.CV},
      url={https://arxiv.org/abs/2104.00650}, 
}

@misc{gao2017talltemporalactivitylocalization,
      title={TALL: Temporal Activity Localization via Language Query}, 
      author={Jiyang Gao and Chen Sun and Zhenheng Yang and Ram Nevatia},
      year={2017},
      eprint={1705.02101},
      archivePrefix={arXiv},
      primaryClass={cs.CV},
      url={https://arxiv.org/abs/1705.02101}, 
}

@misc{wang2023holoassistegocentrichumaninteraction,
      title={HoloAssist: an Egocentric Human Interaction Dataset for Interactive AI Assistants in the Real World}, 
      author={Xin Wang and Taein Kwon and Mahdi Rad and Bowen Pan and Ishani Chakraborty and Sean Andrist and Dan Bohus and Ashley Feniello and Bugra Tekin and Felipe Vieira Frujeri and Neel Joshi and Marc Pollefeys},
      year={2023},
      eprint={2309.17024},
      archivePrefix={arXiv},
      primaryClass={cs.CV},
      url={https://arxiv.org/abs/2309.17024}, 
}

@misc{hendricks2017localizingmomentsvideonatural,
      title={Localizing Moments in Video with Natural Language}, 
      author={Lisa Anne Hendricks and Oliver Wang and Eli Shechtman and Josef Sivic and Trevor Darrell and Bryan Russell},
      year={2017},
      eprint={1708.01641},
      archivePrefix={arXiv},
      primaryClass={cs.CV},
      url={https://arxiv.org/abs/1708.01641}, 
}

@misc{epstein2019oopspredictingunintentionalaction,
      title={Oops! Predicting Unintentional Action in Video}, 
      author={Dave Epstein and Boyuan Chen and Carl Vondrick},
      year={2019},
      eprint={1911.11206},
      archivePrefix={arXiv},
      primaryClass={cs.CV},
      url={https://arxiv.org/abs/1911.11206}, 
}

@misc{grauman2024egoexo4dunderstandingskilledhuman,
      title={Ego-Exo4D: Understanding Skilled Human Activity from First- and Third-Person Perspectives}, 
      author={Kristen Grauman and Andrew Westbury and Lorenzo Torresani and Kris Kitani and Jitendra Malik and Triantafyllos Afouras and Kumar Ashutosh and Vijay Baiyya and Siddhant Bansal and Bikram Boote and Eugene Byrne and Zach Chavis and Joya Chen and Feng Cheng and Fu-Jen Chu and Sean Crane and Avijit Dasgupta and Jing Dong and Maria Escobar and Cristhian Forigua and Abrham Gebreselasie and Sanjay Haresh and Jing Huang and Md Mohaiminul Islam and Suyog Jain and Rawal Khirodkar and Devansh Kukreja and Kevin J Liang and Jia-Wei Liu and Sagnik Majumder and Yongsen Mao and Miguel Martin and Effrosyni Mavroudi and Tushar Nagarajan and Francesco Ragusa and Santhosh Kumar Ramakrishnan and Luigi Seminara and Arjun Somayazulu and Yale Song and Shan Su and Zihui Xue and Edward Zhang and Jinxu Zhang and Angela Castillo and Changan Chen and Xinzhu Fu and Ryosuke Furuta and Cristina Gonzalez and Prince Gupta and Jiabo Hu and Yifei Huang and Yiming Huang and Weslie Khoo and Anush Kumar and Robert Kuo and Sach Lakhavani and Miao Liu and Mi Luo and Zhengyi Luo and Brighid Meredith and Austin Miller and Oluwatumininu Oguntola and Xiaqing Pan and Penny Peng and Shraman Pramanick and Merey Ramazanova and Fiona Ryan and Wei Shan and Kiran Somasundaram and Chenan Song and Audrey Southerland and Masatoshi Tateno and Huiyu Wang and Yuchen Wang and Takuma Yagi and Mingfei Yan and Xitong Yang and Zecheng Yu and Shengxin Cindy Zha and Chen Zhao and Ziwei Zhao and Zhifan Zhu and Jeff Zhuo and Pablo Arbelaez and Gedas Bertasius and David Crandall and Dima Damen and Jakob Engel and Giovanni Maria Farinella and Antonino Furnari and Bernard Ghanem and Judy Hoffman and C. V. Jawahar and Richard Newcombe and Hyun Soo Park and James M. Rehg and Yoichi Sato and Manolis Savva and Jianbo Shi and Mike Zheng Shou and Michael Wray},
      year={2024},
      eprint={2311.18259},
      archivePrefix={arXiv},
      primaryClass={cs.CV},
      url={https://arxiv.org/abs/2311.18259}, 
}

@InProceedings{Heilbron_2015_CVPR,
author = {Caba Heilbron, Fabian and Escorcia, Victor and Ghanem, Bernard and Carlos Niebles, Juan},
title = {ActivityNet: A Large-Scale Video Benchmark for Human Activity Understanding},
booktitle = {Proceedings of the IEEE Conference on Computer Vision and Pattern Recognition (CVPR)},
month = {June},
year = {2015}
}

@inproceedings{seo2020urvos,
  title={Urvos: Unified referring video object segmentation network with a large-scale benchmark},
  author={Seo, Seonguk and Lee, Joon-Young and Han, Bohyung},
  booktitle={European conference on computer vision},
  pages={208--223},
  year={2020},
  organization={Springer}
}

@article{schulman2017proximal,
  title={Proximal policy optimization algorithms},
  author={Schulman, John and Wolski, Filip and Dhariwal, Prafulla and Radford, Alec and Klimov, Oleg},
  journal={arXiv preprint arXiv:1707.06347},
  year={2017}
}

@article{shao2024deepseekmath,
  title={Deepseekmath: Pushing the limits of mathematical reasoning in open language models},
  author={Shao, Zhihong and Wang, Peiyi and Zhu, Qihao and Xu, Runxin and Song, Junxiao and Bi, Xiao and Zhang, Haowei and Zhang, Mingchuan and Li, YK and Wu, Yang and others},
  journal={arXiv preprint arXiv:2402.03300},
  year={2024}
}

@inproceedings{
feng2025groupingroup,
title={Group-in-Group Policy Optimization for {LLM} Agent Training},
author={Lang Feng and Zhenghai Xue and Tingcong Liu and Bo An},
booktitle={The Thirty-ninth Annual Conference on Neural Information Processing Systems},
year={2025},
url={https://openreview.net/forum?id=QXEhBMNrCW}
}

@article{li2026streampro,
  title={StreamPro: From Reactive Perception to Proactive Decision-Making in Streaming Video},
  author={Li, Ao and Xiao, Zihan and Yue, Zihao and Xu, Boshen and Yao, Linli and Li, Jiaze and Fu, Pei and Ju, Jianzhong and Luan, Jian and Jin, Qin},
  journal={arXiv preprint arXiv:2605.16381},
  year={2026}
}

@inproceedings{guan2026video,
  title={Video streaming thinking: Videollms can watch and think simultaneously},
  author={Guan, Yiran and Yin, Liang and Liang, Dingkang and Ju, Jianzhong and Luo, Zhenbo and Luan, Jian and Liu, Yuliang and Bai, Xiang},
  booktitle={European Conference on Computer Vision},
  pages={80--99},
  year={2026},
  organization={Springer}
}

@article{clark2026molmo2,
  title={Molmo2: Open weights and data for vision-language models with video understanding and grounding, 2026},
  author={Clark, Christopher and Zhang, Jieyu and Ma, Zixian and Park, Jae Sung and Salehi, Mohammadreza and Tripathi, Rohun and Lee, Sangho and Ren, Zhongzheng and Kim, Chris Dongjoo and Yang, Yinuo and others},
  journal={URL https://arxiv. org/abs/2601.10611},
  year={2026}
}

@misc{openai2026gpt6astra,
  author = {{OpenAI}},
  title  = {{GPT-6 Astra System Card}},
  year   = {2026},
  month  = sep,
  url    = {https://deploymentsafety.openai.com/gpt-6-astra}
}

@misc{openai2026gpt56,
  author = {{OpenAI}},
  title  = {{GPT-5.6 System Card}},
  year   = {2026},
  month  = jul,
  url    = {https://deploymentsafety.openai.com/gpt-5-6}
}

@misc{qwen3.5,
    title  = {{Qwen3.5}: Towards Native Multimodal Agents},
    author = {{Qwen Team}},
    month  = {February},
    year   = {2026},
    url    = {https://qwen.ai/blog?id=qwen3.5}
}

@article{zhu2025internvl3,
  title={Internvl3: Exploring advanced training and test-time recipes for open-source multimodal models},
  author={Zhu, Jinguo and Wang, Weiyun and Chen, Zhe and Liu, Zhaoyang and Ye, Shenglong and Gu, Lixin and Tian, Hao and Duan, Yuchen and Su, Weijie and Shao, Jie and others},
  journal={arXiv preprint arXiv:2504.10479},
  year={2025}
}

@misc{googledeepmind2026gemini35flash,
  author = {{Google DeepMind}},
  title  = {{Gemini 3.5 Flash Model Card}},
  year   = {2026},
  month  = may,
  url    = {https://storage.googleapis.com/deepmind-media/Model-Cards/Gemini-3-5-Flash-Model-Card.pdf}
}

@inproceedings{zhang2025flash,
  title={Flash-vstream: Efficient real-time understanding for long video streams},
  author={Zhang, Haoji and Wang, Yiqin and Tang, Yansong and Liu, Yong and Feng, Jiashi and Jin, Xiaojie},
  booktitle={2025 IEEE/CVF International Conference on Computer Vision (ICCV)},
  pages={21059--21069},
  year={2025},
  organization={IEEE}
}

@article{driess2023palm,
  title={Palm-e: An embodied multimodal language model},
  author={Driess, Danny and Xia, Fei and Sajjadi, Mehdi SM and Lynch, Corey and Chowdhery, Aakanksha and Ichter, Brian and Wahid, Ayzaan and Tompson, Jonathan and Vuong, Quan and Yu, Tianhe and others},
  journal={arXiv preprint arXiv:2303.03378},
  year={2023}
}

@article{huang2022inner,
  title={Inner monologue: Embodied reasoning through planning with language models},
  author={Huang, Wenlong and Xia, Fei and Xiao, Ted and Chan, Harris and Liang, Jacky and Florence, Pete and Zeng, Andy and Tompson, Jonathan and Mordatch, Igor and Chebotar, Yevgen and others},
  journal={arXiv preprint arXiv:2207.05608},
  year={2022}
}

@inproceedings{
zeng2025streamforest,
title={StreamForest: Efficient Online Video Understanding with Persistent Event Memory},
author={Xiangyu Zeng and Kefan Qiu and Qingyu Zhang and Xinhao Li and Jing Wang and Jiaxin Li and Ziang Yan and Kun Tian and Meng Tian and Xinhai Zhao and Yi Wang and Limin Wang},
booktitle={The Thirty-ninth Annual Conference on Neural Information Processing Systems},
year={2025},
url={https://openreview.net/forum?id=9loSPaBwGO}
}

@InProceedings{Chen_2025_CVPR,
    author    = {Chen, Joya and Zeng, Ziyun and Lin, Yiqi and Li, Wei and Ma, Zejun and Shou, Mike Zheng},
    title     = {LiveCC: Learning Video LLM with Streaming Speech Transcription at Scale},
    booktitle = {Proceedings of the IEEE/CVF Conference on Computer Vision and Pattern Recognition (CVPR)},
    month     = {June},
    year      = {2025},
    pages     = {29083-29095}
}

@article{wu2024videollm,
  title={Videollm-mod: Efficient video-language streaming with mixture-of-depths vision computation},
  author={Wu, Shiwei and Chen, Joya and Lin, Kevin Qinghong and Wang, Qimeng and Gao, Yan and Xu, Qianli and Xu, Tong and Hu, Yao and Chen, Enhong and Shou, Mike Zheng},
  journal={Advances in Neural Information Processing Systems},
  volume={37},
  pages={109922--109947},
  year={2024}
}

@article{li2025videochat,
  title={Videochat: Chat-centric video understanding},
  author={Li, KunChang and He, Yinan and Wang, Yi and Li, Yizhuo and Wang, Wenhai and Luo, Ping and Wang, Yali and Wang, Limin and Qiao, Yu},
  journal={Science China Information Sciences},
  volume={68},
  number={10},
  pages={200102},
  year={2025},
  publisher={Springer}
}

@inproceedings{lin2024video,
  title={Video-llava: Learning united visual representation by alignment before projection},
  author={Lin, Bin and Ye, Yang and Zhu, Bin and Cui, Jiaxi and Ning, Munan and Jin, Peng and Yuan, Li},
  booktitle={Proceedings of the 2024 conference on empirical methods in natural language processing},
  pages={5971--5984},
  year={2024}
}

@article{liu2024streamchat,
  title={Streamchat: Chatting with streaming video},
  author={Liu, Jihao and Yu, Zhiding and Lan, Shiyi and Wang, Shihao and Fang, Rongyao and Kautz, Jan and Li, Hongsheng and Alvare, Jose M},
  journal={arXiv preprint arXiv:2412.08646},
  year={2024}
}

@inproceedings{huang2025online,
  title={Online video understanding: Ovbench and videochat-online},
  author={Huang, Zhenpeng and Li, Xinhao and Li, Jiaqi and Wang, Jing and Zeng, Xiangyu and Liang, Cheng and Wu, Tao and Chen, Xi and Li, Liang and Wang, Limin},
  booktitle={2025 IEEE/CVF Conference on Computer Vision and Pattern Recognition (CVPR)},
  pages={3328--3338},
  year={2025},
  organization={IEEE}
}

@inproceedings{wang2025videollm,
  title={VideoLLM Knows When to Speak: Enhancing Time-Sensitive Video Comprehension with Video-Text Duet Interaction Format.},
  author={Wang, Yueqian and Meng, Xiaojun and Wang, Yuxuan and Liang, Jianxin and Wei, Jiansheng and Zhang, Huishuai and Zhao, Dongyan},
  booktitle={EMNLP (Findings)},
  pages={6338--6359},
  year={2025}
}

@InProceedings{Li_2026_CVPR,
    author    = {Li, Jinzhao and Chen, Yinuo and Piao, Dongxu and Pan, Panwang and Yu, Yifan and Wang, Dong and Yan, Honglei and Yue, Liang and Wang, Shaofei and Chen, Yixin and Huang, Siyuan and Liu, Miao},
    title     = {EgoProx: Evaluating MLLMs on Egocentric 3D Proximity Reasoning Across a Cognitive Hierarchy},
    booktitle = {Proceedings of the IEEE/CVF Conference on Computer Vision and Pattern Recognition (CVPR)},
    month     = {June},
    year      = {2026},
    pages     = {23751-23762}
}
\end{document}